\documentclass{article}

\usepackage{arxiv}

\usepackage[utf8]{inputenc} % allow utf-8 input
\usepackage[T1]{fontenc}    % use 8-bit T1 fonts
\usepackage{hyperref}       % hyperlinks
\usepackage{url}            % simple URL typesetting
\usepackage{booktabs}       % professional-quality tables
\usepackage{amsfonts}       % blackboard math symbols
\usepackage{nicefrac}       % compact symbols for 1/2, etc.
\usepackage{microtype}      % microtypography
\usepackage{lipsum}
\usepackage{lineno,hyperref}
\usepackage{algorithm}
\usepackage{algorithmic}
\usepackage{tabu}
\usepackage{rotating}
\usepackage{graphicx}
\usepackage[lofdepth,lotdepth]{subfig}
\usepackage{tikz}
 
\usepackage{ragged2e}
\usepackage{multirow}
\usepackage{doi}
\usepackage{amsmath}
\hypersetup{allcolors = red}
\usepackage{array}
\usepackage{tablefootnote}
\usepackage{amsmath}
\DeclareMathOperator*{\argsort}{arg\,sort}
\DeclareMathOperator*{\Inception}{Inception}
\modulolinenumbers[5]

\title{A multimodal deep learning approach for named entity recognition from social media}

\author{
  Meysam Asgari-Chenaghlu \thanks{Corresponding author} \\
  Department of Computer Engineering\\
  University of Tabriz\\
  Tabriz, Iran \\
   \texttt{m.asgari@tabrizu.ac.ir} \\
    \And
  M.Reza Feizi-Derakhshi \\
  Department of Computer Engineering\\
  University of Tabriz\\
  Tabriz, Iran \\
  \texttt{mfeizi@tabrizu.ac.ir} \\
   \And
    Leili Farzinvash \\
   Department of Computer Engineering\\
   University of Tabriz\\
   Tabriz, Iran \\
   \texttt{l.farzinvash@tabrizu.ac.ir} \\
   \And
 M. A. Balafar \\
  Department of Computer Engineering\\
  University of Tabriz\\
  Tabriz, Iran \\
  \texttt{balafarila@tabrizu.ac.ir} \\
     \And
 Cina Motamed \\
  Laboratoire d’Informatique Signal et Image de la Côte d’Opale\\
  Université Littoral Côte d’Opale\\
  Calais, France\\
  \texttt{cina.motamed@univ-littoral.fr} \\
}

\begin{document}
\maketitle

\begin{abstract}
Named Entity Recognition (NER) from social media posts is a challenging task. User generated content that forms the nature of social media, is noisy and contains grammatical and linguistic errors. This noisy content makes it much harder for tasks such as named entity recognition. We propose two novel deep learning approaches utilizing multimodal deep learning and Transformers. Both of our approaches use image features from short social media posts to provide better results on the NER task. On the first approach, we extract image features using InceptionV3 and use fusion to combine textual and image features. This presents more reliable name entity recognition when the images related to the entities are provided by the user. On the second approach, we use image features combined with text and feed it into a BERT like Transformer. The experimental results, namely, the precision, recall and F1 score metrics show the superiority of our work compared to other state-of-the-art NER solutions.
\end{abstract}

% keywords can be removed

\section{Introduction}\label{sec:introduction}

A common social media delivery system such as Twitter supports various media types like video, image and text. This media allows users to share their short posts called Tweets. Users are able to share their tweets with other users that are usually following the source user. Hovewer there are rules to protect the privacy of users from unauthorized access to their timeline \cite{Twitter2014}. The very nature of user interactions on Twitter micro-blogging social media is oriented towards their daily life, first witness news-reporting and engaging in various events (sports, political stands, etc.) are examples of such activity. According to studies, news in twitter is propagated and reported faster than conventional news media \cite{Osborne2010}. Thus, extracting first hand news and entities occurring in this fast and versatile online media gives valuable information. However, abridged and noisy content of Tweets makes it even more difficult and challenging for tasks such as named entity recognition and information retrieval \cite{Li2012}.

Named entity recognition task on the scope of social media is very important because many of related tasks directly or indirectly depend on it. This important tool can improve many data analytic and data scientific approaches for different streaming data analysis on various social media platforms. For example, detection of events, hot topics or trending topics from a social media can be done by many methods and systems \cite{atefeh2015survey} while a good named entity recognizer can extract the underlying entities \cite{li2012twiner}. Having the entities and the events at hand, one can easily infer any related information about a person or an entity occurring inside an event.

The task of tracking and recovering information from social media posts is a concise definition of \textit{information retrieval in social media} \cite{Luo2015,Raghavan2004}. However many challenges are blocking useful solutions to this issue, namely, the noisy nature of user generated content and the perplexity of words used in short posts. Sometimes different entities are called the same, for example "Micheal Jordan" refers to a basketball player and also a computer scientist in the field of artificial intelligence. The only thing that divides both of these is the context in which the entity appeared. If the context refers to something related to AI, the reader can conclude "Micheal Jordan" is the scientist, and if the context is refers to sports and basketball then he is the basketball player. The task of distinguishing between different named entities that appear to have the same textual appearance is called \textit{named entity disambiguation}. There is more useful data on the subject rather than on plain text. For example, images and visual data are more descriptive than just text for tasks such as named entity recognition and disambiguation \cite{Davis2012} while some methods only use the textual data \cite{7069115}.

\begin{figure}
	\centering
	\includegraphics[width=0.6\textwidth]{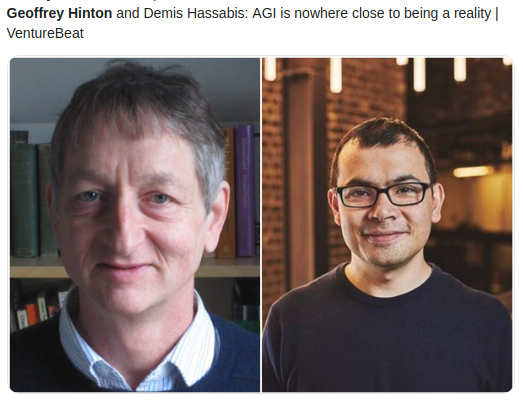}
	\caption{A Tweet containing Image and Text: \textit{Geoffrey Hinton} and \textit{Demis Hassabis}
		are referred in text and respective images are provided with Tweet.}
	\label{fig:1}
\end{figure}

The provided extra information as input is closely related to the textual data. As a clear example, figure \ref{fig:1} shows a tweet containing an image and the related text. The combination of these multimodal data in order to achieve better performance in NLP related tasks is a promising alternative explored recently. Using such well provided data helps the model to have extra features which are useful for NER.

An NLP task such as named entity recognition in social media is a most challenging task because users tend to invent, mistype and epitomize words. Sometimes these words correspond to named entities which makes the recognition task even more difficult \cite{Locke2009}. In some cases, the context that carries the entity (surrounding words and related image) is more descriptive than the entity word presentation \cite{Zhang2018}. This further information, namely, the image related to the post, is more important where it contains no misspells and is straight forward into the topic of the post. In some cases, the context that carries the entity (surrounding words and related image) is more descriptive than the entity word presentation \cite{Zhang2018}.

To find a solution to the issues at hand, and keeping multimodal data in mind, recognition of named entities from social media has become a research interest which utilizes image compared to NER task in a conventional text. Researchers in this field have tried to propose multimodal architectures based on deep neural networks with multimodal input that are capable of combining text and image \cite{Moon2018,Zhang2018, aguilar2017multi}. This multimodal combination can be very misleading when the input image is irrelevant to the text or the textual data in a short sentence form has many out of vocabulary (OoV) tokens. Such inefficiencies are another barrier for the model to provide better results.

In this paper, we draw a better solution in terms of performance by proposing two novel models. One is called CWI (Character-Word-Image model) and the other is based on Transformers. We used multimodal deep neural network combined with Transformers to overcome the NER task in micro-blogging social media. The first model utilizes both of the character and the word level features by using various inner layers. This combination increases the understanding of model over OoV tokens. The skip connection used in character feature extraction when it is combined with other feature extractors, provides more accurate results. We also present use of Transformers in combination with image features. Compared to the base BERT model, our proposed model provides much better results in terms of evaluation metrics. However, these two models are useful in different use-cases; For example in any use-case that the person tag and metrics related to it, is prioritized over other ones, Transformer based model is more accurate; In case of a more general use-case, the first approach is preferred.

The rest of the paper is organized as follows: section \ref{sec:relatedWork} provides an insight view of previous methods; section \ref{sec:propsoedMethod} describes the method we propose; section \ref{sec:experimentalEvaluation} shows experimental evaluation and test results; finally, section \ref{sec:conclusion} concludes the whole article.

\section{Related Work}\label{sec:relatedWork}
Many algorithms and methods have been proposed to detect, classify or extract information from a single type of data such as audio, text, image, etc. However, in the case of social media, data comes in a variety of types such as text, image, video or audio in a bounded style. Most of the time, it is very common to caption a video or image with textual information. This information about the video or image can refer to a person, location and etc. From a multimodal learning perspective, jointly computing such data is considered to be more valuable in terms of representation and evaluation.

Named entity recognition task, on the other hand, is the task of recognizing named entities from a sentence or group of sentences in a document format. Named entity is formally defined as a word or phrase that clearly identifies an item from set of other similar items \cite{Sharnagat2014,6823714}. Equation \ref{eq:1} expresses a sequence of tokens.

\begin{eqnarray}{l}\label{eq:1}
s = \langle w_1, w_2, \dots, w_n \rangle,\\
\label{eq:2}
o = \langle I_s, I_e, t \rangle,\\
\label{eq:3}
o = \langle T_1, T_2, \dots, T_n \rangle.
\end{eqnarray}

From this equation, the NER task is defined as the recognition of tokens that correspond to interesting items. These items from natural language processing perspective are known as named entity categories; BIO2 proposes four major categories, namely, organization, person, location and miscellaneous \cite{sang1999representing}. From the biomedical domain, gene, protein, drug and disease names are known as named entities \cite{6949632,7084590}. The output of NER task is formulated in \ref{eq:2}. $I_s\in [1,N]$ and $I_e\in[1,N]$ is the start and end indices of each named entity and $t$ is named entity type \cite{li2018survey}.

BIO2 tagging for named entity recognition is defined in equation \ref{eq:3}. Table \ref{tab:bio} shows BIO2 tags and their respective meanings; \textit{B} and \textit{I} indicate beginning and inside of the entity respectively, while \textit{O} shows the outside of it. Even though many tagging standards have been proposed for NER task, BIO is the foremost accepted by many real world applications \cite{Manning2014}.

A named entity recognizer gets $s$ as input and provides entity-tags for each token. This sequential process requires information from the whole sentence rather than only tokens and for that reason, it is also considered to be a sequence tagging problem. Another analogous problem to this issue is part of speech tagging and some methods are capable of doing both \cite{huang2015bidirectional}. However, in cases where noise is present and the input sequence has linguistic typos, many methods fail to overcome the problem. As an example, consider a sequence of tokens where a new token invented by social media users gets trended. This trending new word is misspelled and is used in a sequence along with other tokens in which the whole sequence does not follow known linguistic grammar. For this special case, classical methods and those which use engineered features do not perform well. Modern machine learning approaches such as deep learning and character or subword level models perfom better in such problems \cite{stanislawek2019named}.

Using the sequence $s$ itself or adding more information to it divides two approaches: \textit{unimodal} and \textit{multimodal}.
\begin{table}
	\centering
	\caption{BIO Tags and their respective meaning.}\label{tab:bio}
	\begin{tabular}{|m {2 cm}|m {2 cm} |m {4 cm}|}
		\hline
		Begin&End&Description\\ \hline
		B-PER & I-PER&Person\\
		B-LOC & I-LOC&Location \\
		B-ORG & I-ORG&Organization \\
		B-MISC & I-MISC&Miscellaneous \\
		O&O&Outside of entity \\
		\hline
	\end{tabular}
\end{table}
Although many approaches for NER have been proposed and reviewing them is not in the scope of this article, we focus on foremost analogues classical and deep learning approaches for named entity recognition in two subsections. In subsection \ref{sec:relatedWork_1}, unimodal approaches for named entity recognition are presented while in subsection \ref{sec:relatedWork_2}, emerging multimodal solutions are described.

\subsection{Unimodal Named Entity Recognition}\label{sec:relatedWork_1}
The recognition of named entities from only textual data (unimodal learning approach) is a well studied and explored research field. For a prominent example of this category, the Stanford NER is a widely used baseline for many applications \cite{Finkel2005}. The incorporation of non-local information in information extraction is proposed by the authors using Gibbs sampling. The conditional random field (CRF) approach used in this article, creates a chain of cliques, where each clique represents the probabilistic  relationship  between  two adjacent states. Also, the Viterbi algorithm has been used to infer the most likely state in the CRF output sequence. Equation \ref{eq:6} shows the proposed CRF method.

\begin{equation}\label{eq:6}
p(o|s) = \frac{\prod\limits_{i=1}^{n}\phi_i(o_{i-1},o_i,s)}{\sum\limits_{o' \in o}\prod\limits_{i=1}^{n}\phi_i(o'_{i-1},o'_i,s)}
\end{equation}

where $\phi$ is the potential function.

CRF finds the most probable likelihood by modeling the input sequence of tokens $s$ as a normalized product of feature functions. In a simpler explanation, CRF outputs the most probable tags that follow each other. For example, it is more likely to have an \textit{I-PER}, \textit{O} or any other that that starts with \textit{B-} after \textit{B-PER} rather than encountering tags that start with \textit{I-}.

T-NER is another approach that is specifically aimed to answer NER task in twitter \cite{Ritter2011}. A set of algorithms in their original work have been published to answer tasks such as POS (part of speech tagging), named entity segmentation and NER. Labeled LDA has been used by the authors in order to outperform baseline in \cite{Collins1999} for NER task. Their approach strongly relies on the dictionary, contextual and orthographic features.

Deep learning techniques use distributed word or character representation rather than raw one-hot vectors. Most of this research in NLP field use pretrained word embeddings such as \textit{Word2Vec} \cite{mikolov2013efficient}, \textit{GloVe} \cite{Pennington2014} or \textit{fastText} \cite{bojanowski2017enriching}. These low dimensional real valued dense vectors have proved to provide better representation for words compared to one-hot vector or other space vector models.

The combination of word embedding along with bidirectional long-short term memory (LSTM) neural networks are examined in \cite{huang2015bidirectional}. The authors also propose to add a CRF layer at the end of their neural network architecture in order to preserve output tag relativity. Utilization of recurrent neural networks (RNN) provides better sequential modeling over data. However, only using sequential information does not result in major improvements because these networks tend to rely on the most recent tokens. Instead of using RNN, authors used LSTM. The long and short term memory capability of these networks helps them to keep in memory what is important and forget what is not necessary to remember. Equation \ref{eq:f} formulates forget-gate of an LSTM neural network, eq. \ref{eq:i} shows input-gate, eq. \ref{eq:o} notes output-gate and eq. \ref{eq:c} presents memory-cell. Finally, eq. \ref{eq:h} shows the hidden part of an LSTM unit \cite{schuster1997bidirectional,hochreiter1997long}.

\begin{eqnarray}{l}
\label{eq:f}
f_{t}=\sigma _{g}(W_{f}x_{t}+U_{f}h_{t-1}+b_{f}),\\
\label{eq:i}
i_{t}=\sigma _{g}(W_{i}x_{t}+U_{i}h_{t-1}+b_{i}),\\
\label{eq:o}
o_{t}=\sigma _{g}(W_{o}x_{t}+U_{o}h_{t-1}+b_{o}),\\
\label{eq:c}
c_{t}=f_{t}\circ c_{t-1}+i_{t}\circ \sigma _{c}(W_{c}x_{t}+U_{c}h_{t-1}+b_{c}),\\
\label{eq:h}
h_{t}=o_{t}\circ \sigma _{h}(c_{t}).
\end{eqnarray}

For all these equations, $\sigma$ is activation function (\textit{sigmoid} or \textit{tanh} are commonly used for LSTM) and $\circ$ is concatenation operation. $W$ and $U$ are weights and $b$ is the bias which should be learned over training process.

LSTM is useful for capturing the relation of tokens in a forward sequential form; However, in natural language processing tasks, it is required to know the upcoming token. To overcome this problem, the authors have used a backward and forward LSTM combining the output of both.

In a different approach, character embedding followed by a convolution layer is proposed in \cite{ma2016end} for sequence labeling. The utilized architecture is followed by a bidirectional LSTM layer that ends in a CRF layer. Character embedding is a useful technique that the authors tried to use it in a combination with word embedding. Character embedding with the use of convolution as feature extractor from character level, captures relations between characters that form a word and reduces spelling noise. It also helps the model to have an embedding when pretrained word embedding is empty or initialized as random for new words. These words are encountered when they were not present in the training set. Thus, in the test phase, the model fails to provide a useful embedding.

The NLP revolution of "Attention is all you need" was a game changer that eliminated need for any LSTM like sequential methods and replaced it with the scaled dot-product attention and positional encoding in \textbf{Transformer} stacks \cite{vaswani2017attention}. After this new research, many of researchers for various NLP tasks have used the Transformer paradigm; BERT, XLNet, ALBERT and T5 are examples of it \cite{devlin2018bert,yang2019xlnet,lan2019albert,raffel2019exploring}, however, there are many other related works too \cite{minaee2020deep}.

\begin{figure*}[t]
	\centering
	\includegraphics[width=0.4\linewidth]{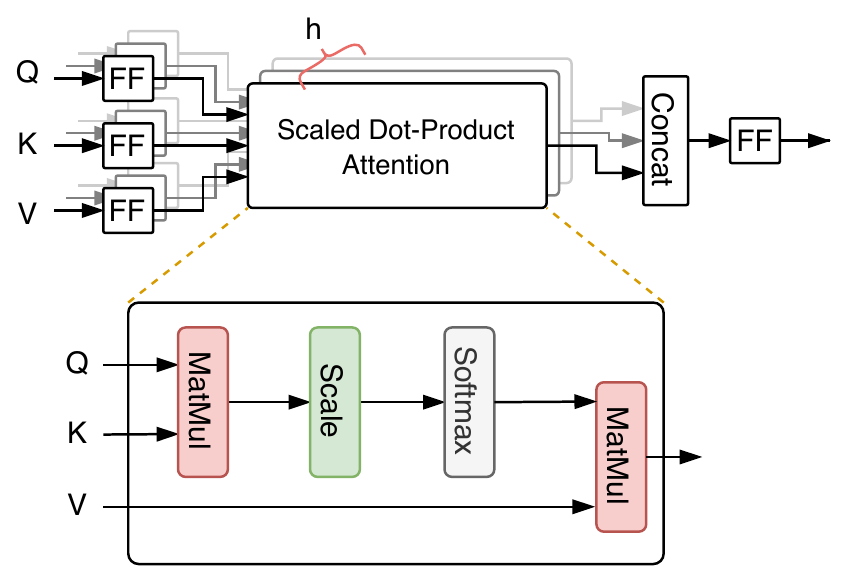}
	\caption{Scaled dot-product attention mechanism.}
	\label{fig:self_attention}
\end{figure*}

The foundation of these methods starts from tokenization and end at training on very huge data with a huge processing power. The tokenization part is done with Byte Pair Encoding (BPE) generally. The idea of utilizing BPE is novel itself in generating tokens even if it was proposed years ago for text compression \cite{shibata1999byte}. The motivation behind using BPE is having better subword parts instead of words or characters \cite{sennrich2015neural}. Figure \ref{fig:self_attention} shows the scaled dot-product attention and the multihead attention mechanism \cite{vaswani2017attention}.

\begin{figure*}[t]
	\centering
	\includegraphics[width=0.5\linewidth]{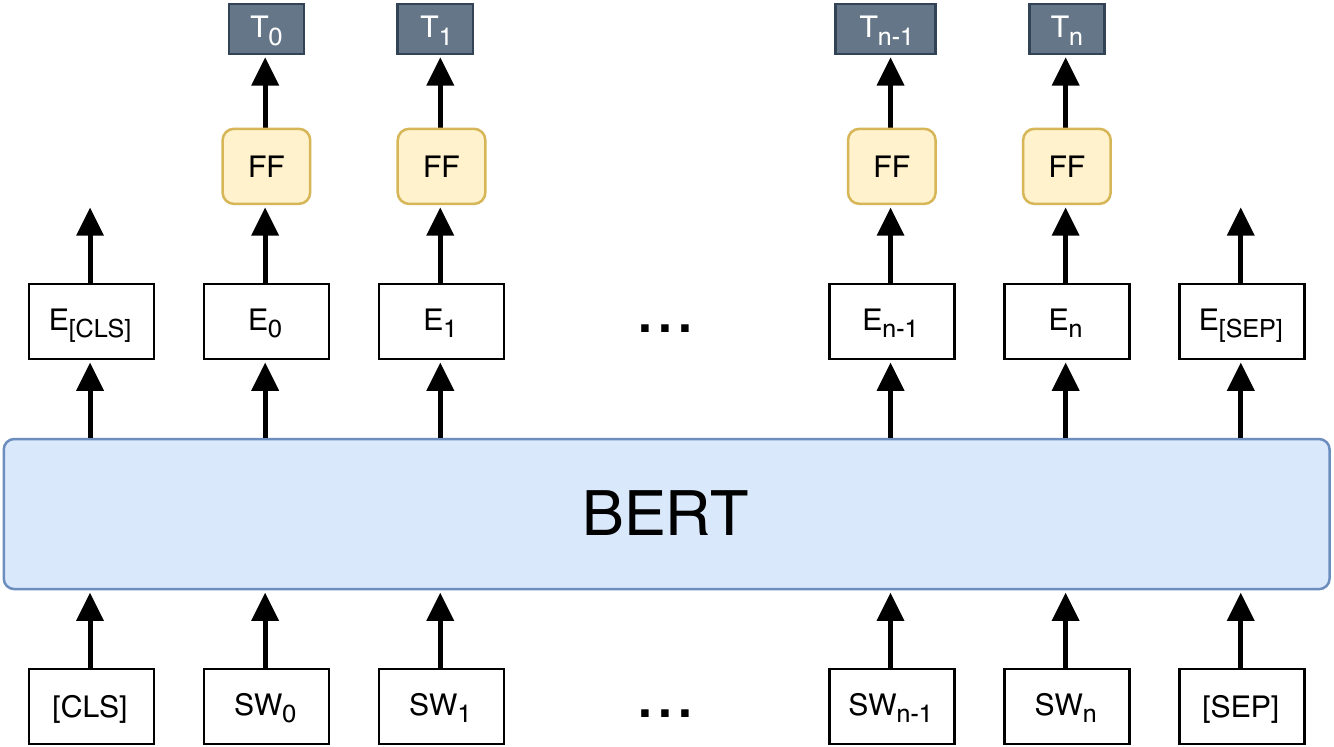}
	\caption{Transformer based NER proposed in \cite{arkhipov2019tuning}.}
	\label{fig:transformer_ner}
\end{figure*}

The attention mechanism has many forms and related studies in fields of machine translation reviewed its effects on the translation task \cite{luong2015effective}. The transformer architecture proposed in \cite{vaswani2017attention} makes use of scaled dot-product attention that is computed using three vectors of Query, Key and Value (Q,K,V). Equation \ref{eq:sdp_att} shows this attention form.

\begin{equation}\label{eq:sdp_att}
Attention(Q,K,V)=softmax(\frac{QK^T}{\sqrt{d_k}})V
\end{equation}
The denominator part of this equation, $\sqrt{d_k}$ is the scale part, proposed in the original article based on the embedding size. The rest of the equation is identical to the figure \ref{fig:self_attention}. Attention head on the other hand, is where scaled dot-product attention units are used in a multi-way, but before using this attention type, a feed-forward (FF as shown in the figure) is applied to each input. A Transformer, is simply a combination of multi-head attention units and feed-forward neural networks. Stacks of Transformer units in encoder and decoder part make a transformer based architecture. However, for many tasks, this architecture is useful. In the case of our study, a typical named entity recognizer architecture based on the Transformer is shown in figure \ref{fig:transformer_ner} \cite{arkhipov2019tuning}. The output embeddings of the last decoder or encoder part is used for generating final NER tags.

\subsection{Multimodal Named Entity Recognition}\label{sec:relatedWork_2}
Multimodal learning has become an emerging research interest and with the rise of deep learning techniques, it has become more visible in different research areas ranging from medical imaging to image segmentation and natural language processing \cite{7976382,7154455,7243358,8052551,7944639,7277066,8336920,7827160,8103116,Moon2018,liu2018learn, beinborn2018multimodal,ngiam2011multimodal,EbrahimiKahou2015,Liu2016,Kahou2016,Suk2014,Cheng2018,Wu2016}. On the other hand, very little research has been focused on the extraction of named entities with joint image and textual data concerning short and noisy content \cite{moon2018multimodal,Esteves2018,Moon2018,Zhang2018} while several studies have been explored in textual named entity recognition using neural models \cite{passos2014lexicon,chiu2016named,huang2015bidirectional,lample2016neural,ma2016end,baldwin2015shared,aguilar2017multi,strubell2017fast}.

State-of-the-art methods have shown acceptable evaluation on structured and well formatted short texts. Techniques based on deep learning such as utilization of convolutional neural networks \cite{strubell2017fast,chiu2016named}, recurrent neural networks \cite{lample2016neural}  and long short term memory neural networks \cite{ma2016end,huang2015bidirectional} are aimed to solve NER problem.

The multimodal named entity recognizers can be categorized in two categories based on the tasks at hand, one tries to improve NER task with the utilization of visual data \cite{moon2018multimodal,Zhang2018,Esteves2018}, and the other tries to give further information about the task at hand such as disambiguation of named entities \cite{Moon2018}. We will refer to both of these tasks as MNER\footnote{Multimodal Named Entity Recognizer}. To have a better understanding of MNER, equation \ref{eq:4} formulates the available multimodal data while equations \ref{eq:2} and \ref{eq:3} are true for this task.

\begin{equation}\label{eq:4}
s' = \langle i, w_1, w_2, \dots, w_n \rangle
\end{equation}

$i$ refers to image and the rest goes same as equation \ref{eq:1} for word token sequence.

In \cite{Esteves2018} pioneering research was conducted using feature extraction from both image and textual data. The extracted features were fed to decision trees in order to output the named entity classes. Researchers have used multiple datasets ranging from buildings to human face images to train their image feature extractor (object detector and k-means clustering) and a text classifier has been trained on texts acquired from \textit{DBPedia}.

Researchers in \cite{moon2018multimodal} proposed a MNER model with regards to triplet embedding of words, characters and image. Modality attention applied to this triplet indicates the importance of each embedding and their impact on the output while reducing the impact of irrelevant modals. Modality attention layer is applied to all embedding vectors for each modal, however the investigation of fine-grained attention mechanism is still unclear \cite{Choi2018}. The proposed method with Inception feature extraction \cite{szegedy2016rethinking} and pretrained \textit{GloVe} word vectors shows good results on the dataset that the authors aggregated from Snapchat\footnote{A multimedia messaging application}. This method shows around 0.5 for precision and F-measure for four entity types (person, location, organization and misc) while for segmentation tasks (distinguishing between a named entity and a non-named entity) it shows around 0.7 for the metrics mentioned.

An adaptive co-attention neural network with four generations are proposed in \cite{Zhang2018}. The adaptive co-attention part is similar to the multimodal attention proposed in \cite{moon2018multimodal} that enabled the authors to have better results over the dataset they collected from Twitter. In their main proposal, convolutional layers are used for word representation, BiLSTM is utilized to combine word and character embeddings and an attention layer combines the best of the triplet (word, character and image features). VGG-Net16 \cite{simonyan2014very} is used as a feature extractor for the image while the impact of other deep image feature extractors on the proposed solution is unclear, however the results show its superiority over related unimodal methods. 

\section{The Proposed Approach} \label{sec:propsoedMethod}
In the present work, we propose two different approaches for the NER problem. First we propose the CWI in subsection \ref{sec:cwi} and in subsection \ref{sec:tran} we demonstrate our second approach, the multimodal transformer. CWI is based on character-word-image features extracted using deep neural network and the multimodal transformer is utilizing transformer combined with image features. Both of these two use the same set of inputs, sentence and the related image from social media posts. For the transformer approach, we used a BERT like transformer that gets the image features extracted using InceptionV3.

\subsection{CWI: Character-Word-Image}
\label{sec:cwi}
\begin{figure*}[t]
	\centering
	\includegraphics[width=0.9\linewidth]{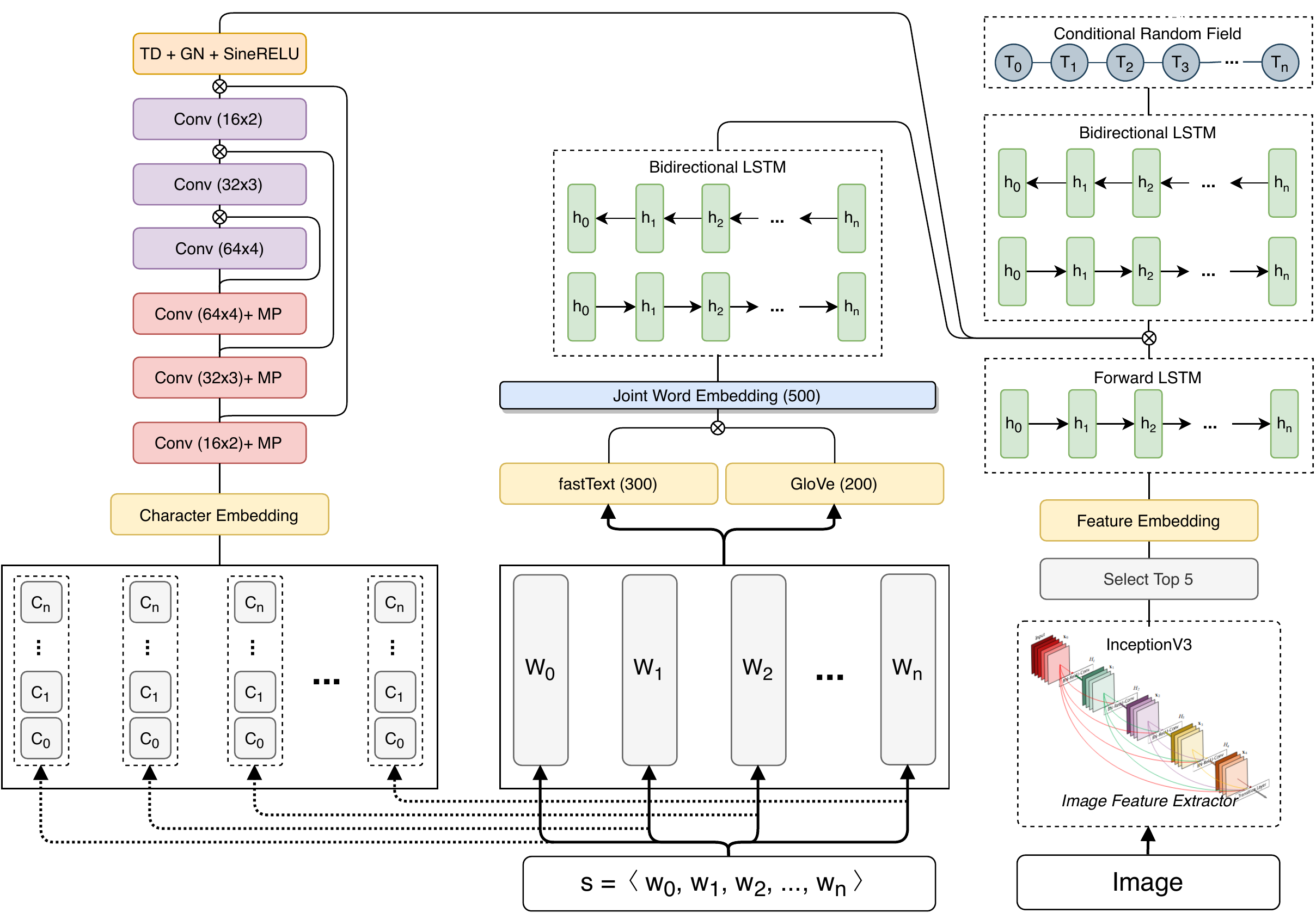}
	\caption{Proposed CWI Model: Character (left), Word (middle) and Image (right) feature extractors combined by bidirectional long-short term memory and the conditional random field at the end.}
	\label{fig:2}
\end{figure*}

In the present work, we propose multimodal deep approach (\textit{CWI}) that is able to handle noise by co-learning semantics from three modalities, character, word and image. Our method is composed of three parts, convolutional character embedding, joint word embedding (fastText-GloVe) and InceptionV3 image feature extraction \cite{szegedy2016rethinking,bojanowski2017enriching,Pennington2014}. Figure \ref{fig:2} shows the \textit{CWI} architecture in more detail.

\textbf{Character Feature Extraction} shown in the left part of figure \ref{fig:2} is a composition of six layers. Each sequence of words from a single tweet, $\langle w_1, w_2, \dots, w_n \rangle$ is converted to a sequence of character representation $\langle [c_{(0,0)}, c_{(0,1)}, \dots, c_{(0,k)}], \dots, [c_{(n,0)}, c_{(n,1)}, \dots, c_{(n,k)}] \rangle$ and in order to apply one dimensional convolution, it is required to be in a fixed length. $k$ shows the fixed length of the character sequence representing each word. Rather than using the one-hot representation of characters, a randomly initialized (uniform distribution) embedding layer is used. The first three convolution layers are followed by a one dimensional pooling layer. In each layer, kernel size is increased incrementally from 2 to 4 while the number of kernels are doubled starting from 16. Just like the first part, the second segment of this feature extractor uses three layers but with slight changes. Kernel size is reduced starting from 4 to 2 and the number of kernels is halved starting from 64. In this part, $\otimes$ sign shows concatenation operation. \textit{TD + GN + SineRelu} note targeted dropout, group normalization and sine-relu \cite{gomez2018targeted, wu2018group, SineRelu}. These layers prevent the character feature extractor from overfitting. Equation \ref{eq:5} defines SineRelu activation function which is slightly different from Relu.

\begin{equation}\label{eq:5}
SineRelu(x)=\begin{cases}
x&x>0\\
\epsilon(\sin{x}-\cos{x}) & x\leq 0
\end{cases}
\end{equation}

Instead of using zero in the second part of this equation, $\epsilon(\sin{x}-\cos{x})$ has been used for negative inputs, $\epsilon$ is a hyperparameter that controls the amplitude of $\sin{x}-\cos{x}$ wave. This slight change prevents the network from having dead-neurons and unlike Relu, it is differentiable everywhere. On the other hand, it has been proven that using GroupNormalization provides better results than BatchNormalization on various tasks \cite{wu2018group}.

However, the dropout has a major improvement on the neural network as an overfitting prevention technique \cite{srivastava2014dropout}, in our setup the TargtedDropout shows to provide better results. TargetedDropout randomly drops neurons whose output is over a threshold. On the other hand, skip connections presented in model, provide better learning in the character feature extraction part and enables it to learn in better way in terms of evaluation metrics.

\textbf{Word Feature Extraction} is presented in the middle part of figure \ref{fig:2}. Joint embeddings from pretrained word vectors of GloVe\footnote{6 billion tokens with 200 dimensional word vectors, available at: \url{http://nlp.stanford.edu/data/glove.6B.zip}} \cite{Pennington2014} and fastText\footnote{16 billion tokens with 300 dimensional word vectors, available at: \url{https://dl.fbaipublicfiles.com/fasttext/vectors-english/wiki-news-300d-1M.vec.zip}} \cite{bojanowski2017enriching} by concatenation operation results in 500 dimensional word embedding. In order to have forward and backward information for each hidden layer, we used a bidirectional long-short term memory \cite{schuster1997bidirectional,hochreiter1997long}. For the words which were not in the pretrained tokens, we used a random initialization (uniform initialization) between -0.25 and 0.25 at each embedding. The result of this phase is extracted features for each word. FastText provides better embeddings when the GloVe fails, and the reason behind it is the structure of fastText itself which is able to capture morphological semantics using subword embeddings.

\textbf{Image Feature Extraction} is shown in the right part of figure \ref{fig:2}. For this part, we have used InceptionV3\footnote{InceptionV3 pretrained model on ImageNet, available at: \url{https://keras.io/applications/\#inceptionv3}} pretrained on ImageNet \cite{deng2009imagenet}. Many models were available as the first part of image feature extraction, however the main reason we used InceptionV3 as feature extractor backbone is better performance of it on ImageNet and the results obtained by this particular model were slightly better compared to others.

Instead of using the headless version of InceptionV3 for image feature extraction, we have used the full model which outputs the 1000 classes of ImageNet. Each of these classes resembles an item, the set of these items can present a person, location or anything that is identified as a whole. To have better features extracted from the image, we have used an embedding layer. In other words, we looked at the top 5 extracted probabilities as words that is shown in eq. \ref{eq:7}; Based on our assumption, these five words present textual keywords related to the image and combination of these words should provide useful information about the objects in visual data. An LSTM unit has been used to output the final image features. These combined embeddings from the most probable items in the image are the key to have extra information from a social media post.

\begin{equation}
\label{eq:7}
\begin{array}{ll}
IW = \underset{x}{\argsort}\lbrace x | x=\Inception(i)\rbrace[1:5],& x \in [0,1]
\end{array}
\end{equation}

where $IW$ is image-word vector, $x$ is output of InceptionV3 and $i$ is the image. $x$ is in domain of [0,1] and $\sum\limits_{\forall k\in x}k=1$ holds true, while $\sum\limits_{\forall k\in IW}k\leq1$.

\textbf{Multimodal Fusion} in our work is presented as the concatenation of three feature sets extracted from words, characters and images. Unlike previous methods, our original work does not include an the attention layer to remove noisy features. Instead, we stacked LSTM units from word and image feature extractors to have better results. The last layer presented at the top right side of figure \ref{fig:2} shows this part. In our second proposed method, we have used attention layer applied to this triplet. Our proposed attention mechanism is able to detect on which modality to increase or decrease focus. Equations \ref{eq:8}, \ref{eq:9} and \ref{eq:10} show attention mechanism related to the second proposed model.

\begin{eqnarray}{l}
\label{eq:8}
u_{it}=\tanh(W_wh_{it}+b_w)\\
\label{eq:9}
\alpha_{it}=\frac{exp(h_t^\top u_{it})}{\sum\limits_{t}\exp(h_t^\top u_{it})}\\
\label{eq:10}
\beta_i = \sum\limits_{t}\alpha_{it}h_{it}
\end{eqnarray}

\textbf{Conditional Random Field} is the last layer in our setup which forms the final output. The same implementation explained in eq. \ref{eq:6} is used for our method.

\subsection{Multimodal Transformer}
\label{sec:tran}
Transformer mechanism described in section \ref{sec:relatedWork_1} is used here with some modification on the hyper-parameters. Also, we changed the input format of the original BERT model that we describe in the current subsection. We call our modified BERT model as MSB (Multimodal Small BERT). The modified version is smaller than the BERT original model and is the same size as small BERT from original BERT released models.

\textbf{Byte Pair Encoding:} For the tokenization part, we have used BPE tokenizer \cite{sennrich2015neural}. The pretrained subotkens are released in \cite{devlin2018bert}\footnote{\href{https://github.com/google-research/bert}{\url{https://github.com/google-research/bert}}}. Before tokenization, we used preprocessing operations such as URL removal. Removing URLs helps the model to skip the unnecessary operations on the input. The rest of text is given to the model with no changes. However, we further pretrained BERT to fit our task at hand, on the related corpus such as crawled twitter corpus but about the tokenizer, we used same as the released version in the original format.

\begin{figure}
	\centering
	\includegraphics[width=0.6\textwidth]{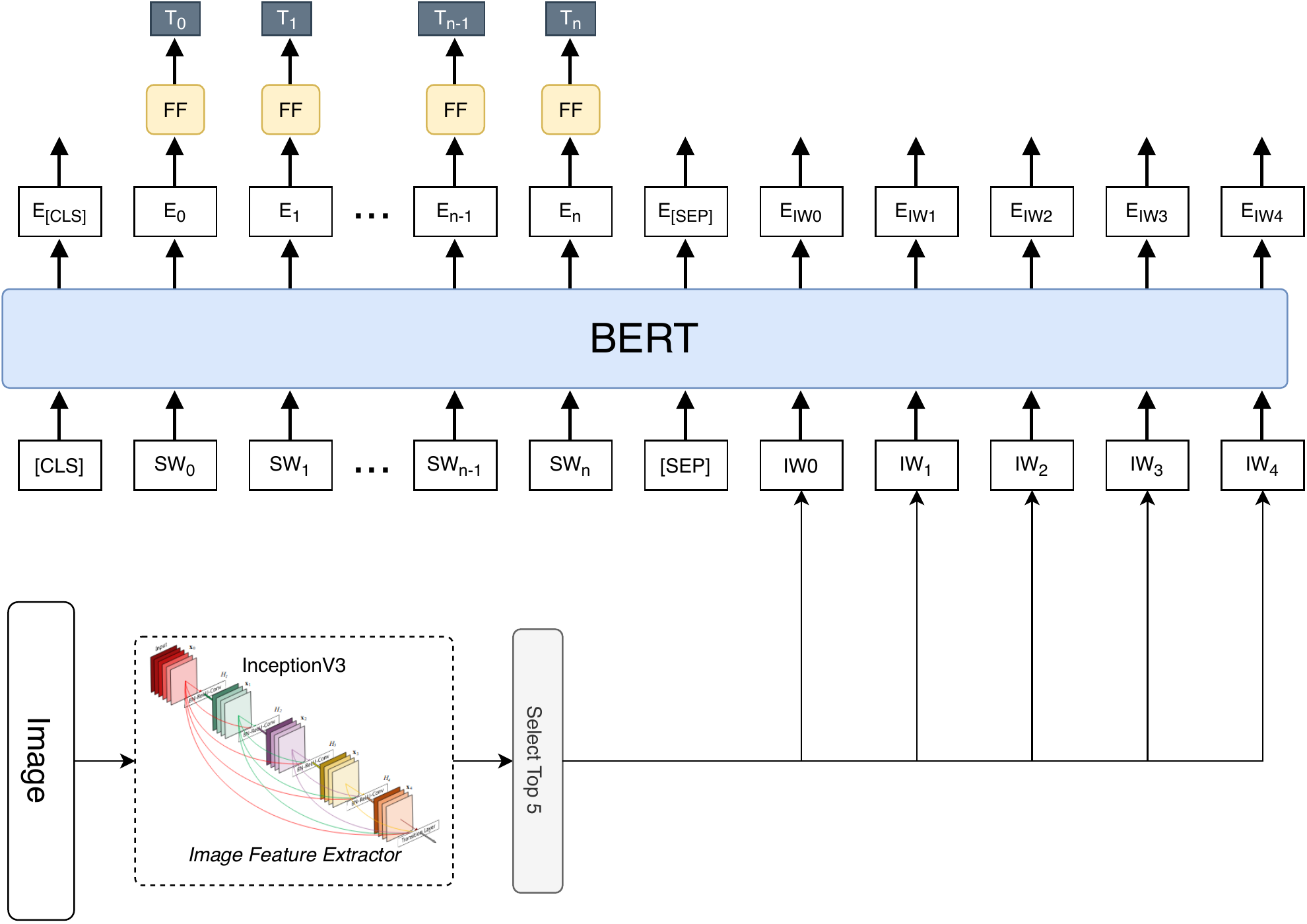}
	\caption{Our proposed second approach: \textbf{M}ultimodal \textbf{S}mall \textbf{B}ERT.}
	\label{fig:bert}
\end{figure}

\textbf{Transformer Configuration:} We used transformers as our building block with getting motivation from BERT as our base and reduced the parameters using the main BERT-Tiny and Small configuration. The configurations we used are presented in table \ref{tab:bert_tiny}. For both of these configurations, the vocabulary size is 30522. Pretrained version are released by google \footnote{BERT-Tiny: \href{https://storage.googleapis.com/bert_models/2020_02_20/uncased_L-2_H-128_A-2.zip}{\url{https://storage.googleapis.com/bert_models/2020_02_20/uncased_L-2_H-128_A-2.zip}}}$^,$\footnote{BERT-Small: \href{https://storage.googleapis.com/bert_models/2020_02_20/uncased_L-4_H-512_A-8.zip}{\url{ https://storage.googleapis.com/bert_models/2020_02_20/uncased_L-4_H-512_A-8.zip}}}. Figure \ref{fig:bert} shows our approach and the utilization of image extracted features into BERT model. The \texttt{[SEP]} token has been used to separate the text and the outputs of image feature extractor (the labels). These two modalities in uniform structure are given to the transformer to extract the final named entities. Another variation of the model is also introduced that has an extra CRF layer. The conditional random field helps the model to correct the mistakes by equation \ref{eq:6}.

Combination of BPE and the transformer gains much improvement in terms of evaluation metrics because it solves the OoV problem and uses a real bidirectional form of NLU instead of left-to-right or right-to-left. This uniform understanding of the splitted tokens with aid of more pretraining on the social media posts and other related details are provided in the next section.

\begin{table}
	\centering
	\caption{Transformer configuration for NER task: Tiny and Small versions.}\label{tab:bert_tiny}
	\begin{tabular}{|l|c|c|c|c|}
		\hline
		Model&Hidden Size&\# of Attention Heads&\# of Transformer Layers\\ \hline
		MSB-Tiny &128&2&2\\
		MSB-Small&512&8&4\\
		\hline
	\end{tabular}
\end{table}

\section{Experimental Evaluation} \label{sec:experimentalEvaluation}
The present section provides evaluation results of our model against baselines. Before diving into our results, a brief description of the dataset and its statistics are provided.

\subsection{Dataset}
In \cite{Zhang2018} a refined collection of tweets gathered from twitter is presented. Their dataset, which is labeled for named entity recognition task contains 8,257 tweets. There are 12,784 entities in total in this dataset. Table \ref{tab:dataset} shows statistics related to each named entity in the training, development and test sets. Following CoNLL-2033 and the BIO2 tagging, this dataset is also tagged manually by experts. Short tweets that contiain less than three words has been discarded by the annotators. Non-English tweets are also discarded. The overall dataset from 26.5 million tweets has been reduced to total 8,257 tweets from 12,784 users. Training, development and testing set is also splitted to 4,000, 1,000 and 3,257 tweets, respectively. All tweets contain images related to them. These images are posted by users and related samples from dataset are presented in 
\ref{fig:samples}.

\begin{table}[h]
	\centering
	\caption{Statistics of named entity types in train, development and test sets \cite{Zhang2018}.}\label{tab:dataset}
	\begin{tabular}{|m {4cm}|c|c|c|c|}
		\hline
		Entity Type&Train&Dev.&Test&Total\\ \hline
		Person &2217&552&1816&4583\\
		Location&2091&522&1697&4308\\
		Organization&928&247&839&2012\\
		Miscellaneous&940&225&726&1881\\ \hline
		Total Entities&6176&1546&5078&12784\\
		\hline
	\end{tabular}
\end{table}

\begin{table*}[h]
	\centering
	\caption{Evaluation results of different approaches compared to ours.}\label{tab:eval}
	\begin{tabular}{|r|l|l|l|l|l|l|l|}
		\hline
		\multirow{2}{*}{Method}							&\multirow{2}{*}{Per.}&\multirow{2}{*}{Loc.}&\multirow{2}{*}{Org.}&
		\multirow{2}{*}{Misc.}&\multicolumn{3}{c|}{Overall}\\\cline{6-8}
		&     &     &     &		&Prec.&Recall&F1\\\hline
		Stanford NER \cite{Finkel2005}				&73.85 &69.35&41.81&21.80&60.98&62.00&61.48\\\hline
		BiLSTM+CRF \cite{huang2015bidirectional}	&76.77&72.56&41.33&26.80&68.14&61.09&64.42\\\hline
		LSTM+CNN+CRF \cite{ma2016end}				&80.86&75.39&47.77&32.61&66.24&68.09&67.15\\\hline
		T-NER \cite{Ritter2011}						&83.64&76.18&50.26&34.56&69.54&68.65&69.09\\\hline
		BiLSTM+CNN+Co-Attention \cite{Zhang2018}	&81.89&\textbf{78.95}&\textbf{53.07}&34.02&72.75&68.74&70.69\\\hline\hline
		CWI (Ours)									&85.81&76.68&50.18&35.65&73.64&69.68&71.61\\\hline
		CWI + Attention (Ours)						&84.02&77.34&52.60&33.47&72.37&70.05&71.19\\\hline
        MSB-Tiny (Ours)						&82.17&76.47&51.09&34.31&71.08&69.75&70.41\\\hline
         MSB-Small (Ours)						&86.32&74.36&50.73&35.12&72.89&70.10&72.74\\\hline
         MSB-Tiny + CRF (Ours)						&84.21&75.16&52.89&35.31&72.87&69.41&71.10\\\hline
          MSB-Small + CRF(Ours)						&\textbf{86.44}&77.16&52.91&\textbf{36.05}&\textbf{74.97}&\textbf{72.04}&\textbf{73.47}\\\hline
	\end{tabular}
\end{table*}

\subsection{Experimental Setup}
In order to obtain the best results in tab. \ref{tab:eval} for our first model (CWI), we have used the following setup in tables \ref{tab:c}, \ref{tab:w}, \ref{tab:i} and \ref{tab:f}. For the second proposed method, the same parameter settings have been used with an additional attention layer. This additional layer has been added after layer 31 in table \ref{tab:f} and before the final CRF layer, indexed as 32. $Adam$ optimizer with $8\times10^{-5}$ has been used in training phase with 10 epochs. The MSB model has been also pretrained on twitter data by using the twitter API and gathered texts. This model has another variation that utilizes the CRF at then last layer for better performance.

\begin{table}
	\centering
	\caption{Implementation details of our model (CWI): Character Feature Extractor.\hspace{\textwidth} \tiny{KS: Kernel Size; PS: Pooling Size; DR: Dropout Rate; TR: Target Rate; $\uparrow$: prior layer\hspace{\textwidth} $\star$ MaxPooling has been applied to second dimension rather than channels}}\label{tab:c}
	\begin{tabular}{|r|r|c|m{3.5cm}|}
		\hline
		ID					&Layer Name			&		Con.		&Details				\\ \hline
		1					&Input 				&		--			&$35\times 40$			\\ \hline
		2					&Embedding			&		$\uparrow$	&Embedding vector size is set to 40 and initialized in range of [-0.25, 0.25] with uniform distribution						\\ \hline
		3					&1D conv.    		&		$\uparrow$	&KS: 2, $\#$ of Kernels: 16\\ \hline
		4					&1D MaxPooling 		&		$\uparrow$	&PS: 2		\\ \hline
		5					&1D conv.    		&		$\uparrow$	&KS: 3, $\#$ of Kernels: 32\\ \hline
		6					&1D MaxPooling 		&		$\uparrow$	&PS: 2		\\ \hline
		7					&1D conv.    		&		$\uparrow$	&KS: 4, $\#$ of Kernels: 64\\ \hline
		8					&1D MaxPooling 		&		$\uparrow$	&PS: 2		\\ \hline
		9					&1D conv.    		&		$\uparrow$	&KS: 4, $\#$ of Kernels: 64\\ \hline
		10					&Concatenation 		&		8,9			&		--				\\ \hline
		11					&1D conv.    		&		$\uparrow$	&KS: 3, $\#$ of Kernels: 32\\ \hline
		12					&Concatenation 		&		6,11		&		--				\\ \hline
		13					&1D conv.    		&		$\uparrow$	&KS: 2, $\#$ of Kernels: 16\\ \hline
		14					&Concatenation 		&		4,13		&		--				\\ \hline
		15					&Targeted Dropout					&		$\uparrow$	&DR: 0.25, TR: 0.4						\\ \hline
		16					&Sine Relu    		&		$\uparrow$	&$\epsilon$: 0.0025						\\ \hline
		17					&Group Normalization			 		&		$\uparrow$	&Applied to 16 groups						\\ \hline
		
	\end{tabular}
\end{table}

\begin{table}
	\centering
	\caption{Implementation details of our model (CWI): Word Feature Extractor.}\label{tab:w}
	\begin{tabular}{|r|r|c|m{4cm}|}
		\hline
		ID					&Layer Name			&		Con.		&Details				\\ \hline
		18					&Input 				&		--			&35						\\ \hline
		19					&GloVe	    		&		18			&GloVe Embedding vector, vector size: 200						\\ \hline
		20					&fastText	 		&		18			&fastText Embedding vector, vector size: 300						\\ \hline
		21					&Concatenation		&		19,20		&--						\\ \hline
		22					&LSTM (Forward)		&		21			&Size: 100						\\ \hline
		23					&LSTM (Backward)	&		21			&Size: 100						\\ \hline
		24					&Concatenation		&		22,23		&--						\\ \hline
	\end{tabular}
\end{table}

\begin{table}
	\centering
	\caption{Implementation details of our model (CWI): Image Feature Extractor.}\label{tab:i}
	\begin{tabular}{|r|r|c|m{4cm}|}
		\hline
		ID					&Layer Name			&		Con.		&Details				\\ \hline
		25					&Input 				&		--			&5 highest probability classes selected from InceptionV3						\\ \hline
		26					&Embedding    		&		$\uparrow$	&50						\\ \hline
		27					&LSTM (Forward) 	&		$\uparrow$	&Size: 50						\\ \hline
	\end{tabular}
\end{table}

\begin{table}
	\centering
	\caption{Implementation details of our model (CWI): Multimodal Fusion.}\label{tab:f}
	\begin{tabular}{|r|r|c|m{4cm}|}
		\hline
		ID					&Layer Name			&		Con.		&Details				\\ \hline
		28					&Concatenation		&		17,24,27	&	--					\\ \hline
		29					&LSTM (Forward) 	&		28			&Size: 100						\\ \hline
		30					&LSTM (Backward) 	&		28			&Size: 100						\\ \hline
		31					&Concatenation		&		29,30		&--						\\ \hline
		32					&CRF			 	&		30			&$\#$ of output Classes: 9, according to BIO2 \ref{tab:bio}						\\ \hline
	\end{tabular}
\end{table}

\begin{table}[ht]
	\centering
	\caption{Effect of different word embedding sizes on our proposed model.}\label{tab:wordvector}
	\begin{tabular}{|r|r|c|c|c|c|}
		\hline
		Embedding Size		&Overall	&Per.	&Loc. 		& Org. 		& Misc.			\\ \hline
		350					&69.87		&81.08	&72.60 		& 49.41 	& 34.03			\\ \hline
		400					&70.23		&83.12	&74.37 		& \textbf{51.87} 	& 34.29			\\ \hline
		500					&\textbf{71.61}		&\textbf{85.81}	&\textbf{76.68}		&50.18		& 35.65			\\\hline
		600					&70.98		&84.29	&75.92 		& 47.81 	& \textbf{38.71}			\\ \hline
	\end{tabular}
\end{table}

For the MSB-Tiny and Small version, we used pretrained weights from BERT that google released. We also trained the model on two different datasets, Twitter-Multimodal-NER dataset (TMN) \cite{Zhang2018} and CoNLL-2003 \cite{sang2003introduction}. The language model that has been used is also trained on the twitter text data that gains more realistic texts to add to the modeling in the pretraining phase. The fine-tuning part has been done in two phases, we first fine-tuned masked language modeling on CoNLL-2003 NER dataset and in the second phase, we trained the whole model on NER task TMN dataset.

\begin{figure}
	\centering
	\includegraphics[width=0.9\textwidth]{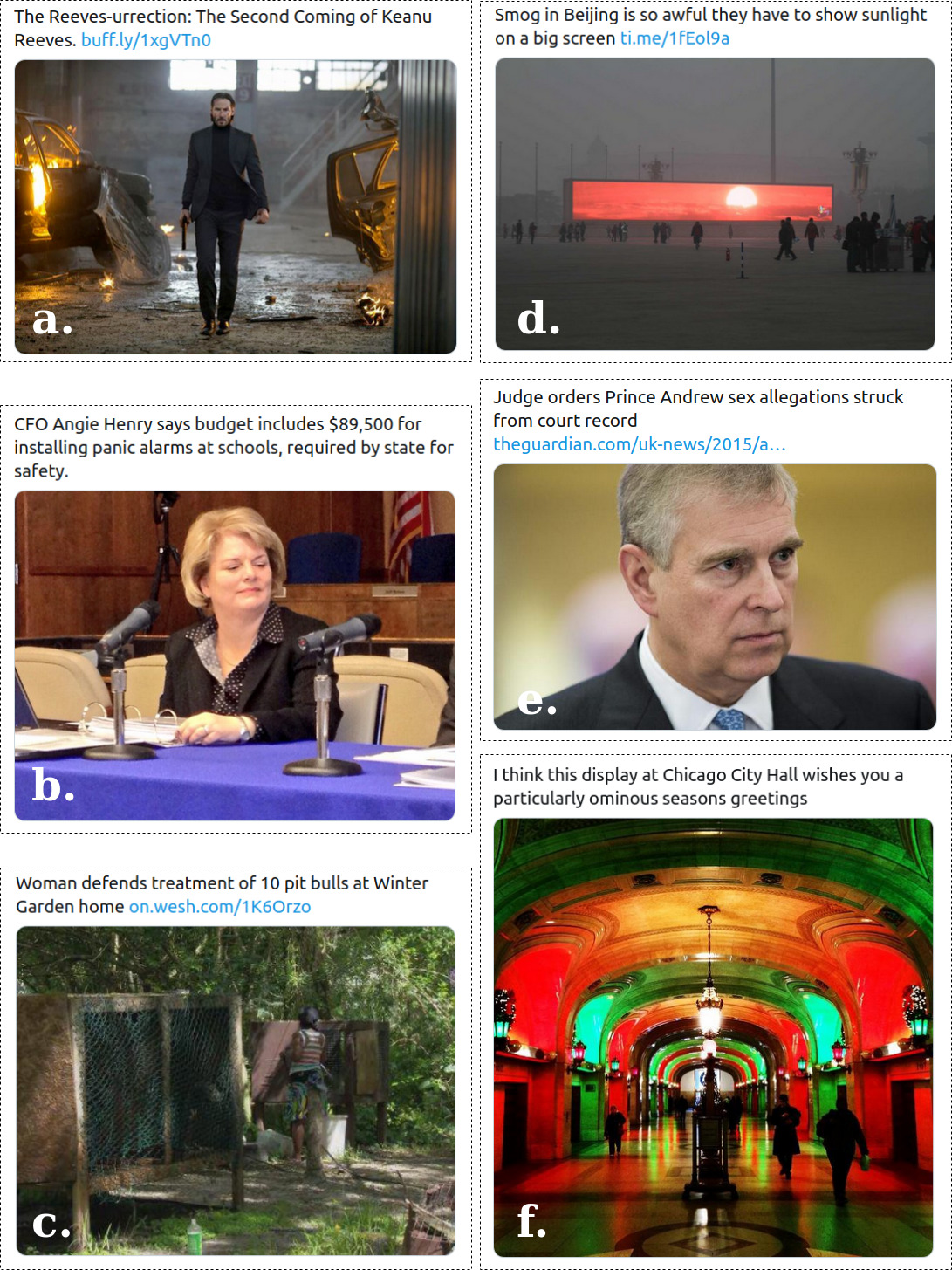}
	\caption{Samples from twitter dataset with text and related image \cite{Zhang2018}.}
	\label{fig:samples}
\end{figure}

\begin{figure}
	\centering
	\includegraphics[width=0.95\textwidth]{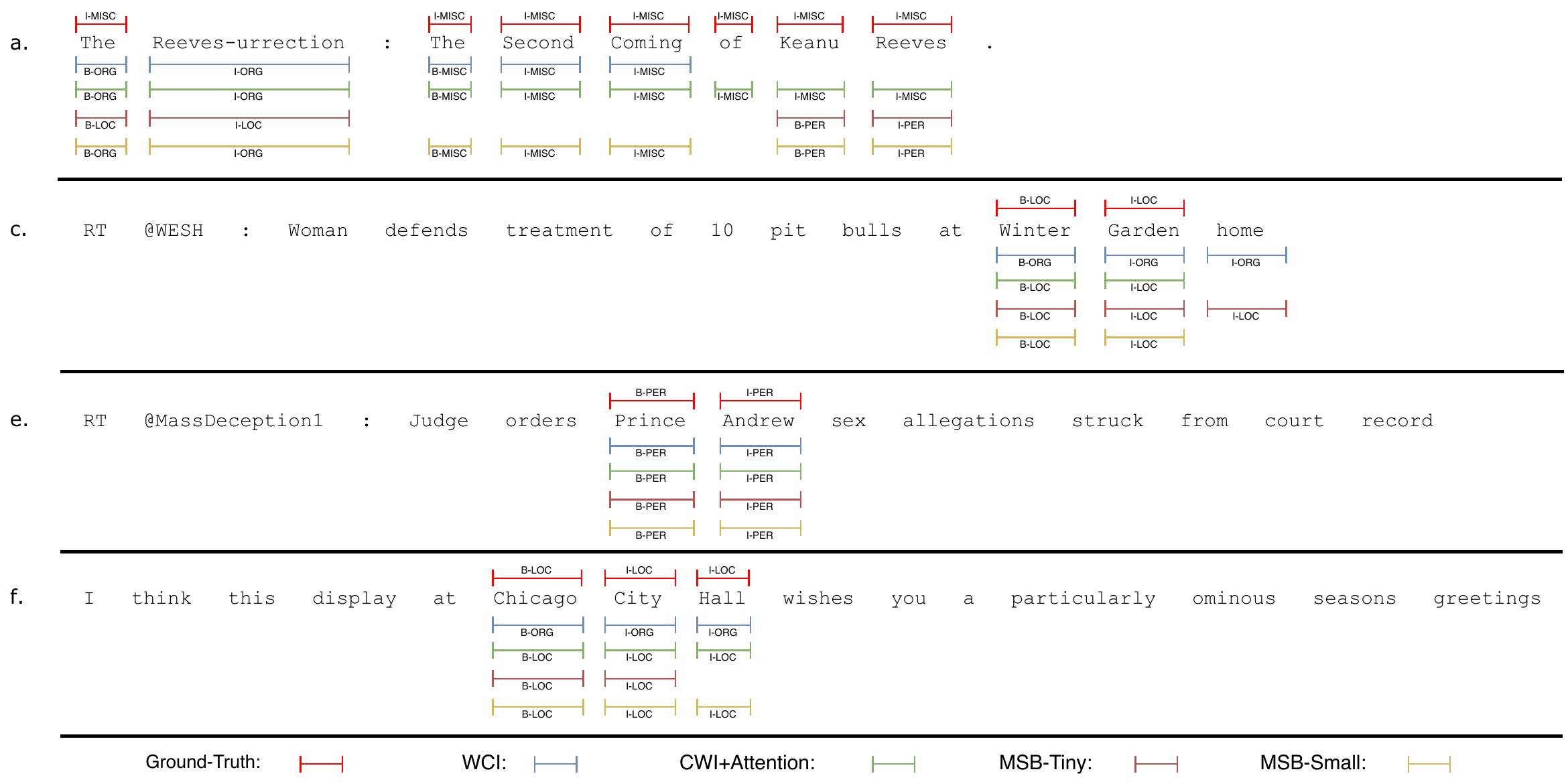}
	\caption{Results of our Approaches on different samples from fig. \ref{fig:samples}, CRF variation of models has been used here.}
	\label{fig:res_samples}
\end{figure}

Figure \ref{fig:samples} shows some visual samples of the dataset. Also, we present the result of our different approaches on these samples in fig. \ref{fig:res_samples}. In this figure, the Ground-Truth is highlighted with red color at the above line of each sentence and the results of our approaches are shown by different colors at the below lines of each sentence. Some samples such as the first one are not correctly labeled in the dataset, but our approach appropriately predicted the true labels.


\subsection{Evaluation Results}
Table \ref{tab:eval} presents the evaluation results of our proposed models. Compared to other state of the art methods, our first proposed model shows $1\%$ improvement on f1 score. The effect of different word embedding sizes on our proposed method is presented in \ref{tab:wordvector}. Sensitivity to TD+SineRelu+GN is presented in tab. \ref{tab:tg}. Second approach is also gains around $3\%$ improvement on the f1 score which is much higher than the first approach but in terms of model size and training time it has downsides. The training time for the first model in a CoreI7 processor with Nvidia GeForce GTX 1650 is around 10 minutes while for the second approach it is 2 days for pretraining on twitter data and fine tuning on the datasets. The training time for the InceptionV3 is not considered because we used the original released version with no changes from Keras Applications.

\begin{table}[ht]
	\centering
	\caption{Effect of \textit{TD+GN+SineRelu} on our proposed model.}\label{tab:tg}
	\begin{tabular}{|r|r|c|c|c|c|}
		\hline
		TD+GN+SineRelu		&Overall	&Per.	&Loc. 		& Org. 		& Misc.			\\ \hline
		No					&64.18		&76.21	&72.30 		&40.98 	& 28.81			\\ \hline
		Yes					&\textbf{71.61}		&\textbf{85.81}	&\textbf{76.68}		&\textbf{50.18}		& \textbf{35.65}			\\\hline
	\end{tabular}
\end{table}

\section{Conclusion} \label{sec:conclusion}
In this article, we have proposed two NER approaches based on multimodal deep learning. In our first model, we have used a new architecture in character feature extraction that has helped our model to overcome the issue of noise. We also used transformers as our building block to propose MSB. Instead of using direct image features from near last layers of image feature extractors such as Inception, we have used the direct output of the last layer. The last layer is 1000 classes of diverse objects that is result of InceptionV3 trained on ImageNet dataset. We used top 5 classes out of these and converted them to one-hot vectors. The resulting image feature embedding out of these high probability one-hot vectors helped our model to overcome the issue of noise in images posted by social media users. Evaluation results of our proposed models compared to other state of the art methods show its superiority to these methods in overall while in two categories (Person and Miscellaneous) our model outperformed others.

\section{Acknowledgments} \label{sec:akc}
The authors would like to thank Shervin Minaee for reviewing this work, and providing very useful comments to improve this work.
%\section*{References}

\bibliographystyle{unsrt}  

\bibliography{bib}

\end{document}